\pdfoutput=1
\documentclass[11pt]{article}

\usepackage{EMNLP2022}

\usepackage{times}
\usepackage{latexsym}
\usepackage{amsfonts}
\usepackage{graphicx}
\usepackage{url} 
\usepackage{amsmath}
\usepackage{ulem}
\usepackage{booktabs}
\usepackage{multirow}
\usepackage{makecell}

\usepackage[T1]{fontenc}
\usepackage[utf8]{inputenc}

\usepackage{microtype}

\usepackage{inconsolata}

\title{RaP: Redundancy-aware Video-language Pre-training for Text-Video Retrieval}

\author{
    Xing Wu\textsuperscript{\rm 1,2,3},Chaochen Gao\textsuperscript{\rm 1,2}\thanks{The first two authors contribute equally.}, Zijia Lin\textsuperscript{\rm 3},Zhongyuan Wang\textsuperscript{\rm 3},Jizhong Han\textsuperscript{\rm 1},Songlin Hu\textsuperscript{\rm 1,2}\thanks{Corresponding author.}
    \\
    \textsuperscript{\rm 1}Institute of Information Engineering, Chinese Academy of Sciences\\
    \textsuperscript{\rm 2}School of Cyber Security, University of Chinese Academy of Sciences\\
    \textsuperscript{\rm 3}Kuaishou Technology
    \\
    \{gaochaochen,zangliangjun,hanjizhong,husonglin\}@iie.ac.cn
    \\\{wuxing,wangzhongyuan\}@kuaishou.com, linzijia07@tsinghua.org.cn
}

\begin{document}
\maketitle
\begin{abstract}
Video language pre-training methods have mainly adopted sparse sampling techniques to alleviate the temporal redundancy of videos.
Though effective, sparse sampling still suffers inter-modal redundancy: visual redundancy and textual redundancy. Compared with highly generalized text, sparsely sampled frames usually contain text-independent portions, called visual redundancy. Sparse sampling is also likely to miss important frames corresponding to some text portions, resulting in textual redundancy. Inter-modal redundancy leads to a mismatch of video and text information, hindering the model from better learning the shared semantics across modalities. 
To alleviate it, we propose Redundancy-aware Video-language Pre-training. 
We design a redundancy measurement of video patches and text tokens by calculating the cross-modal minimum dis-similarity. 
Then, we penalize the high-redundant video patches and text tokens through a proposed redundancy-aware contrastive learning.
We evaluate our method on four benchmark datasets, MSRVTT, MSVD, DiDeMo, and LSMDC, achieving a significant improvement over the previous state-of-the-art results. Our code are available at \href{https://github.com/caskcsg/VLP/tree/main/RaP}{https://github.com/caskcsg/VLP/tree/main/RaP}.
\end{abstract}

\section{Introduction}
% With the rapid development of online video applications, text-to-video cross-modal retrieval (text-video retrieval for short) has become increasingly important in recent years.
Text-video retrieval computes the semantic similarity between a text query and candidate videos, ranking more similar videos higher.
Video-language pre-training can jointly learn the representation of video and text, allowing cross-modal similarity computation to be more effective and efficient, so it has been widely explored in text-video retrieval \cite{bain2021frozen, li2022align, lei2021less, gorti2022x}.
Videos are composed of dozens or hundreds of consecutive frames, usually containing much redundant information, already known as temporal redundancy.
% Many previous methods use offline extracted video features to avoid the expensive computational overhead required in training for many frames.
% Video feature extractors are usually trained on task-independent data, and the resulting fixed features are not ideal for VLP and downstream tasks.
% Unlike these methods, 
\cite{lei2021less} proposes to sparsely sample frames from videos to alleviate temporal redundancy without incurring any drop in effect, followed by many works \cite{bain2021frozen, li2022align, gorti2022x}.
% , which does not rely on any video feature extractor and allows end-to-end training of the entire network. 
% Many following works \cite{bain2021frozen, li2022align, gorti2022x} adopt the sparse sampling strategy and achieve good results in text-video retrieval.

\begin{figure}
\includegraphics[width=7.5cm]{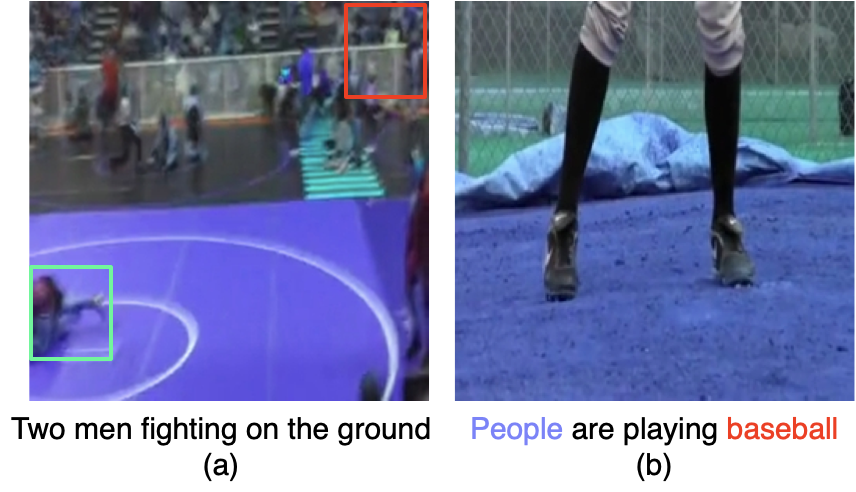}
\setlength{\belowcaptionskip}{-0.5cm}
\caption{Examples of inter-modal redundancy. (a) Visual redundancy: the pixels in the red box are redundant with respect to the text description. (b) Textual redundancy: the token ``baseball'' in red font does not correspond to any portion in the video frame.}
\label{Rap_case}
\end{figure}
\begin{figure*}[!htbp]
\centering
\includegraphics[width=16cm]{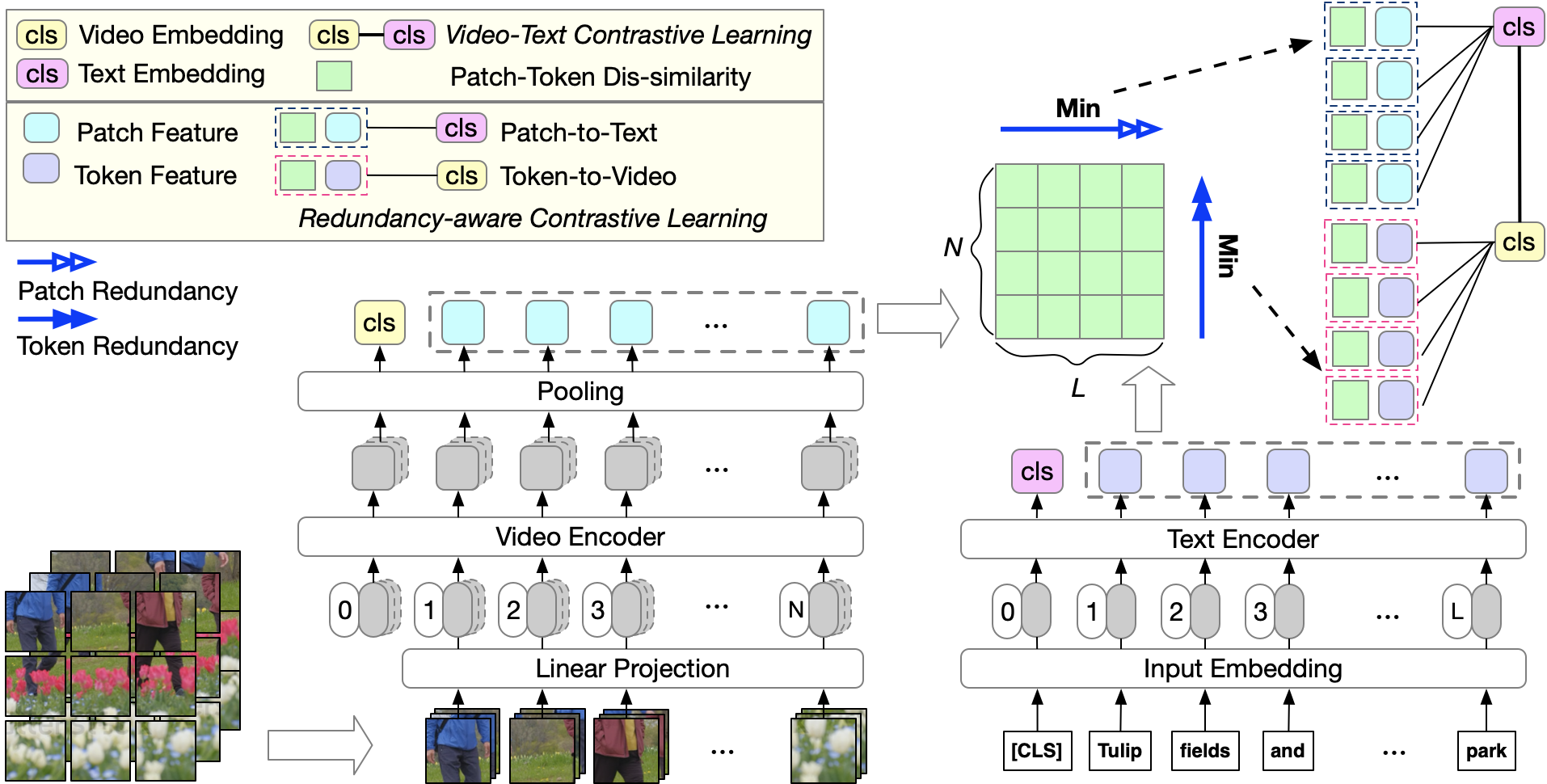}
\setlength{\belowcaptionskip}{-0.5cm}
\caption{Redundancy-aware video-language pre-training method. Sparsely sampled frames are mapped into video embedding and patch features through multiple layers. Similarly, the text is also mapped into text embedding and token features. The dis-similarity matrix between the patch and token features is used to calculate the redundancy. We take the minimum value by row/column as the redundancy of each patch/token, respectively. Patch/token redundancy is then used for weighted patch-to-text/token-to-video contrastive learning to reduce the impact of high-redundancy patches/tokens.}
\label{Rap}
\end{figure*}

In addition to intra-modal redundancy, i.e., temporal redundancy, there is inter-modal redundancy between video and text.
Some previous works \cite{zhu2020actbert, chen2020uniter, wang2022object, li2022align} focus on modeling fine-grained alignment, which can alleviate inter-modal redundancy to some extent. But they have not categorized and analyzed inter-modal redundancy in details.
We summarize inter-modal redundancy into two categories: visual redundancy and textual redundancy, as shown in the example in Figure \ref{Rap_case}.
\textbf{Visual redundancy} refers to the redundant information beyond textual semantics that existed in sparsely sampled frames.
% Specifically, video frames are composed of continuous natural signals, while the text is discretely generalized. 
In contrast to highly generalized text, multiple video frames tend to contain portions that are semantically irrelevant to the text.
% As shown in Figure \ref{Rap_case}-(a), a large amount of background pixels in the frame have little relevance to the text.
\textbf{Textual redundancy} refers to the redundant portions in the text that are irrelevant to sparsely sampled frames. 
Sparsely sampling from the video will probably miss important frames associated with some text portions.
% As shown in Figure \ref{Rap_case}-(b),  the ``baseball'' token does not appear in the frame. 

Inter-modal redundancy will lead to the mismatch of video and text semantics, preventing the model from better learning shared semantics across modalities.
% Visual redundancy is not conducive to learning text semantics. 
Visually redundant pixels are encoded into the video embedding, and pre-training aligns the text embedding with the redundant video embedding, pushing the text embedding away from the correct text semantics.
Similarly, pre-training aligns the video embedding with the redundant text embedding, pushing the video embedding away from the correct video semantics.
Methods to alleviate redundancy through fine-grained alignment \cite{zhu2020actbert, chen2020uniter, wang2022object} mainly rely on offline object detectors to extract objects or tags from sampled frames.
These methods are based on the assumption that the objects extracted from the sampled frames are related to the text description, or that the tags extracted in one frame relate to other frames. 
However, there is uncertainty in the correlation between multiple frames, especially with sparse sampling. 
In addition, object detectors have the drawbacks of inaccurate detection, a limited number of categories, and unable to perform end-to-end optimization in video language pre-training.

To better alleviate the problem of inter-modal redundancy, we first propose a redundancy measurement, as shown in Figure \ref{Rap}.
In video-language pre-training, each video frame is split into patches, and a text is tokenized into tokens.
Take the Figure \ref{Rap_case}-(a) as an example. The green patch is low-redundant because it relates to ``two'', ``men'' and ``fighting'' tokens. In contrast, the red patch is high-redundant because it corresponds to no token.
Therefore, \textbf{the redundancy of a patch depends on how well it corresponds to the tokens in the text}.
In other words, a patch is high-redundant if it has low semantic similarity to all tokens.
So we use the minimum dis-similarity between a patch and all tokens as its visual redundancy
Symmetrically, \textbf{ the redundancy of a token depends on how well it corresponds to the patches in the video}, and we use the minimum dis-similarity between a token and all patches as its textual redundancy.

To reduce the impact of high-redundant patches on learning text embedding or tokens on learning video embedding, we then propose redundancy-aware contrastive learning.
We take patch-text pairs as additional positives in video-to-text constrastive learning and assign smaller weights to pairs with high-redundant patches.
Similarly, we take token-video pairs as additional positives in text-to-video constrastive learning and assign smaller weights to those with high-redundant tokens.
Specially, the weight equals $(1 - redundancy)$ in calculation.

Combining the above two points, we propose \textbf{R}edundancy-\textbf{a}ware Video-language \textbf{P}re-training (RaP) method, which is end-to-end trainable without relying on object detection, as shown in Figure \ref{Rap}. 
We evaluate RaP on four text-video retrieval datasets, MSRVTT, MSVD, DiDeMo, and LSMDC, achieving a significant improvement over the previous state-of-the-art results. Sufficient ablation studies also confirm the effectiveness of RaP:

Our contributions can be summarized as follows:
\begin{enumerate}
    \item We summarize the inter-modal redundancy in video-language pre-training and propose a measurement for the redundancy.
    % We analyze the shortcomings of object-based method to alleviate visual redundancy and propose textual redundancy for the first time.
    \item We propose redundancy-aware contrastive learning to alleviate the two inter-modal redundancy and facilitate high-quality modelling of the shared semantics.
    \item Experimental results show that our method significantly improves the state-of-the-art results on multiple text-video retrieval datasets.
\end{enumerate}

\section{Related Work}

Video-language pre-training aims to learn joint representations between video and language.
Videos consist of consecutive frames and often contain visually similar redundant information. Redundant information will bring two problems to video-language pre-training. One is the extra computational overhead, and the other is that the semantics of video and text cannot be well aligned. 

To alleviate the problem of overhead, prior approaches\cite{li2020hero,luo2020univl,miech2020end,miech2019howto100m,sun2019videobert,zhu2020actbert} use offline tools to extract video features, but cannot achieve end-to-end pre-training. ClipBERT\cite{lei2021less} efficiently trains the video encoder end-to-end using only a few sparsely sampled frames. 
Later, some well-performing methods \cite{bain2021frozen,li2022align, wang2022object,fu2021violet} adopt the sparse sampling strategy to alleviate temporal redundancy to reduce computational overhead.
Since we focus on mitigating misleading caused by redundant information,
% Since we focus on reducing redundancy between video and text, 
rather than temporal redundancy, we follow \cite{lei2021less} to use sparsely sampled frames as input to the video encoder.

To better align video and text, some recent works introduce fine-grained alignment in video-language pre-training  \cite{zhu2020actbert, chen2020uniter, wang2022object, li2022align}. These works identify regions with objects in video via offline trained object detectors or prompters. Then they align regions containing the objects with the text description, which alleviates inter-modal redundancy implicitly.
Unlike them, we explicitly propose an efficient redundancy measure that quantifies the impact of different redundancy.
The most related one is OA-Trans \cite{wang2022object}, which extracts objects and tags from an anchor frame with an offline detector, and uses the object-related regions or tags as additional input to reduce redundancy.
Unlike OA-Trans \cite{wang2022object}, our learnable redundancy measurement matrix can be optimized end-to-end in pre-training.
We do not rely on an offline detector, so we do not suffer the drawbacks of offline detectors. 

\section{Backgroud}
This section introduces some background knowledge of video-language pre-training, including video-text input, encoders, and contrastive learning for training.

\subsection{Text-Video Input}
Video-language pre-training methods use text-video pairs as raw input, where the text is a description of the video. 
A video $\mathcal{V}$ is sparsely sampled $K$ frames, obtaining a sequence of frames $\{\mathcal{F}_k\}_ {k=0}^{K}$ 
Before being fed into the video encoder, frame $\mathcal{F}_k$ will be divided into $N$ same-sized frame patches $\{\mathcal{P}^k_n\}_{n=0}^{ N}$.
Frame patches are then mapped into input embeddings $\{\mathcal{P}e^k_n\}_{n=0}^{ N}$ via projection, where $\mathcal{P}e^k_0$ is an additional [CLS] embedding to learn the global semantics of frame $\mathcal{F}_k$.
Similarly, before being fed into the text encoder, a text $\mathcal{T}$ will be tokenized into $L$ consecutive tokens and projected into token embeddings $\{\mathcal{T}e_l\}_{l=0}^{L}$, where $\mathcal{T}e_0$ is an additional [CLS] embedding to learn the global semantics of text.

\subsection{Text-Video Encoders}
\paragraph{Video Encoder}
We use visual transformer (ViT) \cite{dosovitskiy2020image} as the video encoder to process each frame $\mathcal{F}_k$ separately. ViT takes frame patch embeddings $\{\mathcal{P}e^k_n\}_{n=0}^{ N}$ as input, and output frame patch features $\{\mathcal{P}f^k_n\}_{n=0}^{ N}$ corresponding to the $N+1$ positions .
Then, we perform mean pooling operation on the features of the same position across $K$ frames. We further transform the pooled feature of each position into a shared normalized low-dimensional (e.g. 256-dim) space, obtaining the video patch feature: 
$\mathbf{P}_n = \frac{1}{K} \sum\limits_{k=0}^{K} \mathcal{P}f^k_n$.
Unless otherwise specified, we will refer to the $\mathbf{P}_n$ as \textbf{patch feature} for convenience hereafter. Particularly, $\mathbf{P}_{cls}$ denotes the global \textbf{video embedding} in the [CLS] position.

\paragraph{Text Encoder}
We use BERT \cite{devlin2018bert} as a text encoder to process text, which takes embeddings $\{\mathcal{T}e_l\}_{l=0}^{L}$ as input. The output embedding of each corresponding position of BERT is transformed into the above low-dimensional space as the token feature embedding $\{\mathbf{T}_l\} _{l=0}^{L}$ . Unless otherwise specified, we will refer to the $\mathbf{T}_l$ as \textbf{token feature} for convenience hereafter. Particularly, $\mathbf{T}_{cls}$ denotes the global \textbf{text embedding} in the [CLS] position.

\subsection{Video-Text Contrastive Learning}
Following CLIP\cite{radford2021learning}, we align video and text features into a comparable shared embedding space via contrastive learning. 
Given the normalized video embedding $\mathbf{P}_{cls}$ and normalized text embedding $\mathbf{T}_{cls}$, the similarity function between video V and text T is:
\begin{equation}
\mathbf{s}(\mathbf{P}_{cls}, \mathbf{T}_{cls}) = \mathbf{P}_{cls} \cdot \mathbf{T}_{cls}
\end{equation}
We aim to assign higher similarity scores to matched video-text pairs. Therefore, in contrastive learning, we take matched video-text pairs as positives and all other pairs formed in a batch as negatives. Given a batch with $B$ matched pairs $\{\mathbf{P}_{cls}^i, \mathbf{T}_{cls}^i\}_{i=1}^{B}$, the video-text contrastive loss of each pair $\{\mathbf{P}_{cls}^i, \mathbf{T}_{cls}^i\}$ consists of two symmetric terms, one for video-to-text contrastive learning:
\begin{equation}
\mathcal{L}_{\mathrm{v} 2 \mathrm{t}}=-\log \frac{\exp(\mathbf{s}\left(\mathbf{P}_{cls}^i, \mathbf{T}_{cls}^i\right) / \tau)}{\sum\limits_{j=1}^{B} \exp(\mathbf{s}\left(\mathbf{P}_{cls}^i, \mathbf{T}_{cls}^j\right) / \tau)}
\end{equation}
and the other for text-to-video contrastive learning:
\begin{equation}
\mathcal{L}_{\mathrm{t} 2 \mathrm{v}}=-\log \frac{\exp(\mathbf{s}\left(\mathbf{T}_{cls}^i, \mathbf{P}_{cls}^i\right) / \tau)}{\sum\limits_{j=1}^{B} \exp(\mathbf{s}\left(\mathbf{T}_{cls}^i, \mathbf{P}_{cls}^j\right) / \tau)}
\end{equation}

\section{Redundancy-aware Video-language Pre-training}
In this section, we introduce our proposed Redundancy-aware video-language pre-training in details. An overview of our approach can refer to Figure \ref{Rap}.
First, we introduce how to measure cross-modal redundancy. Then, we introduce how to reduce the impact of redundancy on video-language pre-training.

\subsection{Cross-modal Minimum Dis-similarity as Redundancy}
We use the similarity function in equation (1) to calculate the cross-modal dis-similarity between a patch feature $\mathbf{P}_n$ and a token feature $\mathbf{T}_l$:
\begin{equation}
\mathbf{d}(\mathbf{P}_n, \mathbf{T}_l) = 1 - \mathbf{s}(\mathbf{P}_n, \mathbf{T}_l)
\end{equation}
As shown in Figure \ref{Rap}, we calculate the dis-similarity between all \textit{non}-[CLS] patch features $\{\mathbf{P}_n\}_{n=1}^{ N}$ and all \textit{non}-[CLS] token features $\{\mathbf{T}_l\} _{l=1}^{L}$, resulting in a dis-similarity matrix $\mathbb{M}$ with dimension $N \times L$. 
Each row of $\mathbb{M}$ denotes the dis-similarities between a patch feature $\mathbf{P}_n$ and all \textit{non}-[CLS] token features $\{\mathbf{T}_l\} _{l=1}^{L}$. 
We take the minimum value of $\mathbb{M}$ by the row as the visual redundancy of each patch:
\begin{equation}
\mathbf{vr}_n = \mathbf{min}(\{\mathbb{M}_{nl}\}_{l=1}^{L})
\end{equation}
Symmetrically, each column of $\mathbb{M}$ denotes the dis-similarities between a token feature $\mathbf{T}_l$ and all \textit{non}-[CLS] patch features $\{\mathbf{P}_n\} _{n=1}^{N}$. 
We take the minimum value of $\mathbb{M}$ by the column as the textual redundancy of each token:
\begin{equation}
\mathbf{tr}_l = \mathbf{min}(\{\mathbb{M}_{nl}\}_{n=1}^{N})
\end{equation}

\subsection{Redundancy-aware Video-Text Contrastive Learning}
\label{method}
We use the redundancy of patches and tokens to improve the contrastive learning process. 
\paragraph{Redundancy-aware video-to-text contrastive learning} In the original video-to-text contrastive learning, there is only one positive text sample for a given video. To reduce the impact of textual redundancy on video embedding learning, we treat all \textit{non}-[CLS] tokens in the positive text as positives too, but we assign higher weights to low-redundancy token features in the loss calculation:
\begin{equation}
\mathcal{L}^{\mathrm{R}}_{\mathrm{v} 2 \mathrm{t}}=-\log \frac{ \sum\limits_{l=1}^{L} \mathbf{w}_l \cdot \exp(\mathbf{s}\left(\mathbf{P}_{cls}^i, \mathbf{T}_l^i\right) / \tau)}{\sum\limits_{j=1}^{B} \sum\limits_{l=1}^{L} \exp(\mathbf{s}\left(\mathbf{P}_{cls}^i, \mathbf{T}_{l}^j\right) / \tau)}
\end{equation}
, where the $\mathbf{w}_l = 1 - \mathbf{tr}_l$. The weighted loss constrains video embeddings to pay more attention to low-redundancy token features while ignoring the high-redundancy ones.

\paragraph{Redundancy-aware text-to-video contrastive learning} Symmetrically, in the original text-to-video contrastive learning, there is only one positive video sample for a given text. To reduce the impact of visual redundancy on text embedding learning, we treat all \textit{non}-[CLS] video patches in the positive video as positives too, but we assign higher weights to low-redundancy patch features in the loss calculation:
\begin{align}
&\mathcal{L}^{\mathrm{R}}_{\mathrm{t} 2 \mathrm{v}}=-\log \frac{\sum\limits_{n=1}^{N} \mathbf{w}_n \cdot \exp(\mathbf{s}\left(\mathbf{T}_{cls}^i, \mathbf{P}_n^i\right) / \tau)}{\sum\limits_{j=1}^{B}\sum\limits_{n=1}^{N} \exp(\mathbf{s}\left(\mathbf{T}_{cls}^i, \mathbf{P}_{n}^j\right) / \tau)}
\end{align}
, where the $\mathbf{w}_n = 1 - \mathbf{vr}_n$. The weighted loss constrains text embeddings to pay more attention to low-redundancy patch features while ignoring the high-redundancy ones.

\paragraph{Redundancy-aware contrastive learning} Overall, redundancy-aware contrastive learning (RaCL) in both directions constrains the embeddings of one modality to focus more on the low-redundant local features of the other modality.
Therefore, RaCL allows different modalities to guide mutually to learn the correct shared semantics. The RaCL loss is defined as the sum of losses in both directions:
\begin{align}
&\mathcal{L}_{RaCL}=\mathcal{L}^{\mathrm{R}}_{\mathrm{v} 2 \mathrm{t}}+\mathcal{L}^{\mathrm{R}}_{\mathrm{t} 2 \mathrm{v}}
\end{align}
Video-text pre-training usually trains some auxiliary tasks to help convergence, such as LM tasks, The losses of these tasks are collectively referred to as $\mathcal{L}_{\mathrm{others}}$. So the total loss is calculated as:
\begin{align}
&\mathcal{L}_{}=\mathcal{L}_{\mathrm{others}}+\lambda*\mathcal{L}_{RaCL} \label{loss}
% \subsection{Backbone Network}
\end{align}
, where $\lambda$ is a balance hyperprarameter. 

\section{Experiments}
We conduct experiments on the four most commonly used text-video retrieval benchmark datasets, which will introduced in the \ref{sec:Downstreamtask} section. Following existing literature \cite{li2022align,bain2021frozen}, we report Recall@1 (R1), Recall@5 (R5), Recall@10 (R10) and Median Rank (MdR).

\subsection{Backbone Network}
We use the \cite{li2022blip} network as the backbone network for our video-language pre-training, with a ViT video encoder and a BERT text encoder.
Since this is not the core of this paper, we leave the detailed network structure and auxiliary tasks to the appendix \ref{sec:appendix}. 
Considering that our method is an optimization of video-text contrastive learning, our method can also be effective on other backbone networks using contrastive learning.

\subsection{Pre-training data}
Following the recent works \cite{li2022align,bain2021frozen,fu2021violet}, we jointly pre-train RaP on image-text and video-text datasets, which we briefly describe below. \\
\textbf{WebVid2.5M(WebVid2M)}\cite{bain2021frozen} contains 2.5M image and text pairs collected from the web. The text data in WebVid describes the global video semantics.\\
\textbf{Google Conceptual Captions (CC3M)}\cite{Sharma2018ConceptualCA} consists of 3.3M image-text pairs from the web. \\
During pre-training, we make static videos by duplicate images from CC3M. Thus our pre-training data contains 5.5M video-text pairs, fewer than the widely used HT100M dataset\cite{miech2019howto100m}.
Note that, for a fair comparison, we do not compare with the works \cite{luo2021clip4clip,gorti2022x,cheng2021improving} initialized from CLIP\cite{radford2021learning}, which has already been pre-trained on over 400M image-text pairs.

\subsection{Text-Video Retrieval Datasets}
\label{sec:Downstreamtask}
\textbf{MSRVTT}\cite{xu2016msr} consists of 10K videos, each paired with about 20 human-labeled captions. We train with 7k/9k videos and report results for 1k test split.\\
\textbf{DiDeMo}\cite{anne2017localizing} contains 10K Flickr videos, annotated with 40K sentences. Following \cite{lei2021less,luo2020univl,liu2019use,li2022align}, we evaluate paragraph-to-video retrieval, where all sentence descriptions of videos are concatenated into a single query. For a fair comparison with previous methods, we do not use the ground-truth proposals for temporal localization.\\
\textbf{MSVD}\cite{chen2011collecting} contains 1,970 videos from YouTube and 80k English descriptions.Training, validation, and test splits consist of 1,200, 100, and 670 videos, respectively.\\
\textbf{LSMDC}\cite{rohrbach2015dataset} is a clip dataset of 118,081 videos, each with a caption description. The length of the video varies from 2 seconds to 30 seconds. In LSMDC, the training split consists of 101,079 videos, the validation split consists of 7,408 videos. We report results on the test split which contains 1,000 videos.

\begin{table}[!tbp]
\centering
\scalebox{0.7}{
\begin{tabular}{l|cccc}
\toprule 
\textbf{Method} & \textbf{R1}$\uparrow$ & \textbf{R5}$\uparrow$& \textbf{R10}$\uparrow$ & \textbf{MdR}$\downarrow$\\
\midrule
\midrule
\multicolumn{5}{c}{Zero-shot} \\
\midrule
% HT100M\cite{miech2019howto100m}$\ddagger$ & 7.5 & 21.2 & 29.6 & 38.0 \\
ActBERT\cite{zhu2020actbert}$\ddagger$ & 8.6 & 23.4 & 33.1 & 36.0 \\
MIL-NCE\cite{miech2020end}$\ddagger$ & 9.9 & 24.0 & 32.4 & 29.5 \\
SupportSet\cite{patrick2020support}$\ddagger$ & 8.7 & 23.0 & 31.1 & 36.0 \\
VideoClip\cite{xu2021videoclip}$\ddagger$ & 10.4 & 22.2 & 30.0 & - \\
FiT\cite{bain2021frozen}$\clubsuit$ & 18.7 & 39.5 & 51.6 & 10.0 \\
VIOLET\cite{fu2021violet}$\clubsuit$ & 25.9 & \textbf{49.5} & \textbf{59.7} & - \\
% DemoVLP\cite{cai2022revitalize}$\clubsuit$ & 21.7 & 42.2 & 52.2 & - \\
HD-VILA\cite{xue2022advancing}$\diamondsuit$ & 14.4 & 31.6 & 41.6 & 17.5 \\
OA-Trans\cite{wang2022object}$\clubsuit$ & 23.4 & 47.5 & 55.6 & 8.0 \\

ALPRO\cite{li2022align}$\clubsuit$  & 24.1 & 44.7 & 55.4 & 8.0 \\
\midrule
RaP(ours)$\clubsuit$ & \textbf{28.9} & 47.5 & 56.8 & \textbf{7.0} \\
\midrule
\midrule
\multicolumn{5}{c}{Fine-tuning on 7k training videos} \\
\midrule
JSFusion\cite{yu2018joint}  & 10.2 & 31.2 & 43.2 & 13.0 \\
HT100M\cite{miech2019howto100m}$\ddagger$ & 14.9 & 40.2 & 52.8 & 9.0 \\
ActBERT\cite{zhu2020actbert}$\ddagger$ & 16.3 & 42.8 & 56.9 & 10.0 \\
HERO\cite{li2020hero}$\ddagger$  & 16.8  & 43.4  & 57.7 & -  \\
AVLNet\cite{le2020hierarchical}$\ddagger$ & 27.1 & 55.6 & 66.6 & 4.0 \\

VideoClip\cite{xu2021videoclip}$\ddagger$ & 30.9 & 55.4 & 66.8 & 4.0 \\ 

NoiseEst\cite{amrani2021noise}$\ddagger$  & 17.4 & 41.6 & 53.6 & 8.0 \\
ClipBERT\cite{lei2021less}  & 22.0  & 46.8  & 59.9 & 6.0 \\

COTS\cite{lu2022cots}$\triangle$ & 32.1 & 60.8 & 70.2 & 3.0 \\
ALPRO\cite{li2022align}$\clubsuit$ & 33.9 & 60.7 & 73.2 & 3.0 \\
\midrule
RaP(ours)$\clubsuit$ & \textbf{38.5} & \textbf{64.0} & \textbf{74.4} & 3.0 \\
\midrule
\midrule
\multicolumn{5}{c}{Fine-tuning on 9k training videos} \\
\midrule
SupportSet\cite{patrick2020support}$\ddagger$ & 30.1 & 58.5 & 69.3 & 3.0 \\
FiT\cite{bain2021frozen}$\clubsuit$ & 31.0 & 59.5 & 70.5 & 3.0 \\

VIOLET\cite{fu2021violet}$\clubsuit$ & 34.5 & 63.0 & 73.4 & - \\
COTS\cite{lu2022cots}$\triangle$ & 36.8 & 63.8 & 73.2 & 2.0 \\
HD-VILA\cite{xue2022advancing}$\diamondsuit$ & 35.0 & 65.2 & \textbf{77.2} & 3.0 \\
OA-Trans\cite{wang2022object}$\clubsuit$ & 35.8 & 63.4 & 76.5 & 3.0 \\

\midrule
RaP(ours)$\clubsuit$ & \textbf{40.9} & \textbf{67.2} & 76.9 & \textbf{2.0} \\
\bottomrule
\end{tabular}}
\caption{Comparisons with existing text-to-video retrieval state-of-the-art methods with zero-shot and fine-tuning setups on MSRVTT. $\clubsuit$: Methods using WebVid2M and CC3M datasets(5.5M). $\ddagger$: Methods using HT100M\cite{miech2019howto100m} dataset. $\diamondsuit$:Methods using HD-VILA-100M\cite{xue2022advancing} dataset. $\triangle$:Methods using 15.3M image-text pairs dataset.}
\label{table_msrvtt}
\end{table}

\subsection{Implementation Details}
During pre-training, we conduct experiments on 64 NVIDIA V100 GPUs using PyTorch framework\cite{paszke2017automatic}. We initialize our video encoder with ViT-B/16\cite{dosovitskiy2020image} with 12 layers. The text encoder is initialized by BERT$_{base}$\cite{devlin2018bert}. We randomly sample 4 frames from each video and resize each frame to $256 \times 256$. Then we split each resized frame into patches. AdamW\cite{loshchilov2017decoupled} is adopt as the optimizer with a weight of 0.05. We train the model for 30 epochs with a batch size of 1920 (30 per GPU). The learning rate is initialized as 1e-6 and warmed to 3e-4 after 3,000 training iterations. We select the final checkpoint to fine-tune text-video retrieval datasets.

During the fine-tuning stage, we perform our experiment on 8 NVIDIA V100 GPUs. We sparsely sample 4 frames and resize them to the same video frame size(256*256) as the pre-training stage. The learning rate is initialized as 1e-5. For each benchmark dataset, we select a checkpoint according to the results of the validation split and inference the checkpoint on the test split. For MSRVTT9k without a validation split, we train the model for 10 epochs and choose the final checkpoint. When inference, following\cite{li2022align}, we uniformly sample 8 frames for each video to ensure reproducibility.

\begin{table}[!tbp]
\centering
\scalebox{0.7}{
\begin{tabular}{l|cccc}
\toprule 
\textbf{Method} & \textbf{R1}$\uparrow$ & \textbf{R5}$\uparrow$& \textbf{R10}$\uparrow$ & \textbf{MdR}$\downarrow$\\
\midrule
\midrule
\multicolumn{5}{c}{Zero-shot} \\
\midrule
VideoClip\cite{xu2021videoclip}  & 16.6 & 46.9 & - & - \\
FiT\cite{bain2021frozen}$\clubsuit$  & 21.1 & 46.0 & 56.2 & 7.0 \\
VIOLET\cite{fu2021violet}$\clubsuit$  & 23.5 & 49.8 & 59.8 & - \\
OA-Trans\cite{wang2022object}$\clubsuit$ & 23.5 & 50.4 & 59.8 & 6.0 \\
ALPRO\cite{li2022align}$\clubsuit$  & 23.8 & 47.3 & 57.9 & 6.0 \\
\midrule
RaP(ours)$\clubsuit$  & \textbf{29.5} & \textbf{55.7} & \textbf{65.6} & \textbf{4.0} \\
\midrule
\midrule
\multicolumn{5}{c}{Fine-tuning} \\
\midrule
% S2VT\cite{venugopalan2014translating} & 11.9 & 33.6 & - & 13.0 \\
% FSE\cite{zhang2018cross} & 13.9 & 36.0 & - & 11.0 \\
% MoEE\cite{miech2018learning}  & 16.1 & 41.2 & 55.2 & 8.0 \\
% CE\cite{liu2019use} & 16.1 & 41.1 & - & 8.0 \\
ClipBERT\cite{lei2021less}  & 20.4  & 48.0  & 60.8 & 6.0 \\
TT-CE\cite{croitoru2021teachtext} & 21.6 & 48.6 & 62.9 & 6.0 \\
FiT\cite{bain2021frozen}$\clubsuit$ & 31.0 & 59.8 & 72.4 & 3.0 \\
VIOLET\cite{fu2021violet}$\clubsuit$  & 32.6 & 62.8 & 74.7 & - \\
ATP\cite{buch2022revisiting} & 26.1 & 50.5 & - & - \\ 
HD-VILA\cite{xue2022advancing}$\diamondsuit$ & 26.0 & 54.8 & 69.0 & 4.0 \\
OA-Trans\cite{wang2022object}$\clubsuit$ & 34.8 & 64.4 & 75.1 & 3.0 \\
ALPRO\cite{li2022align}$\clubsuit$  & 35.9 & 67.5 & 78.8 & 3.0 \\
\midrule
RaP(ours)$\clubsuit$  & \textbf{42.9} & \textbf{71.2} & \textbf{80.2} & \textbf{2.0} \\
\bottomrule
\end{tabular}}
\caption{Comparisons with existing text-to-video retrieval state-of-the-art methods with zero-shot and fine-tuning setups on DiDeMo. $\clubsuit$: Methods using WebVid2M and CC3M datasets(5.5M). $\diamondsuit$:Methods using HD-VILA-100M\cite{xue2022advancing} dataset.}
\label{table_DiDeMo}
\end{table}

\begin{table}[!t]
\centering
\scalebox{0.7}{
\begin{tabular}{l|cccc}
\toprule 
\textbf{Method}  & \textbf{R1}$\uparrow$ & \textbf{R5}$\uparrow$& \textbf{R10}$\uparrow$ & \textbf{MdR}$\downarrow$\\
\midrule
\midrule
\multicolumn{5}{c}{Zero-shot} \\
\midrule
SupportSet\cite{patrick2020support}$\ddagger$  & 21.4 & 46.2 & 57.7 & 6.0 \\
% DemoVLP\cite{cai2022revitalize}$\clubsuit$  & 41.6 & 69.8 & 80.5 & - \\
\midrule
RaP(ours)$\clubsuit$  & \textbf{35.9} & \textbf{64.3} & \textbf{73.7} & \textbf{3.0} \\
\midrule
\midrule
\multicolumn{5}{c}{Fine-tuning} \\
\midrule
VSE\cite{kiros2014unifying} & 12.3 & 30.1 & 42.3 & - \\
VSE++\cite{faghri2017vse++} & 15.4 & 39.6 & 53.0 & 9.0 \\
Multi.Cues\cite{mithun2018learning} & 20.3 & 47.8 & 61.1 & 6.0 \\
CE\cite{liu2019use} & 19.8 & 49.0 & 63.8 & 6.0 \\
SupportSet\cite{patrick2020support}$\ddagger$  & 28.4 & 60.0 & 72.9 & 4.0 \\
FiT\cite{bain2021frozen}$\clubsuit$  & 33.7 & 64.7 & 76.3 & 3.0 \\
OA-Trans\cite{wang2022object}$\clubsuit$ & 39.1 & 68.4 & 80.3 & 2.0 \\
% DemoVLP}\cite{cai2022revitalize}$\clubsuit$$\clubsuit$ & 50.9} & 78.9} & 87.0} & - \\
\midrule
RaP(ours)$\clubsuit$  & \textbf{45.4} & \textbf{74.8} & \textbf{83.6} & \textbf{2.0} \\
\bottomrule
\end{tabular}}
\caption{Comparisons with text-to-video retrieval state-of-the-art methods with zero-shot and fine-tuning setups on MSVD. We treat each sentence as the textual query.  $\clubsuit$: Methods using WebVid2M and CC3M datasets(5.5M).}
\label{table_msvd}
\end{table}

\begin{table}[!t]
\centering
\scalebox{0.7}{
\begin{tabular}{l|cccc}
\toprule 
\textbf{Method}  & \textbf{R1}$\uparrow$ & \textbf{R5}$\uparrow$& \textbf{R10}$\uparrow$ & \textbf{MdR}$\downarrow$\\
\midrule
\midrule
\multicolumn{5}{c}{Zero-shot} \\
\midrule
% DemoVLP$\clubsuit$  & \textbf{14.3} & 25.8 & 32.0 & - \\
RaP(ours)$\clubsuit$  & 12.8 & 26.6 & 33.4 & 37.0 \\
\midrule
\midrule
\multicolumn{5}{c}{Fine-tuning} \\
\midrule
JSFusion\cite{yu2018joint}  & 9.1 & 21.2 & 34.1 & 36.0 \\
MoEE\cite{miech2018learning}  & 9.3 & 25.1 & 33.4 & 27.0 \\
CE\cite{liu2019use}  & 11.2 & 26.9 & 34.8 & 25.3 \\
MMT\cite{gabeur2020multi}  & 12.9 & 29.2 & 38.8 & 19.3 \\
AVLNet\cite{le2020hierarchical}$\ddagger$ & 17.0 & 38.0 & 48.6 & - \\
Dig\cite{ijcai2021-154} & 15.8 & 34.1 & 43.6 & - \\
MDMMT\cite{dzabraev2021mdmmt} & 18.8 & 38.5 & 47.9 & - \\
% DemoVLP}\cite{cai2022revitalize}$\clubsuit$$\clubsuit$  & 25.2} & 45.5} & 54.5} & - \\
FiT\cite{bain2021frozen}$\clubsuit$  & 15.0 & 30.8 & 39.8 & 20.0 \\
VIOLET\cite{fu2021violet}$\clubsuit$  & 16.1 & 36.6 & 41.2 & - \\
VTMCE\cite{9706971} & 14.9 & 33.2 & - & - \\
HD-VILA\cite{xue2022advancing}$\diamondsuit$ & 17.2 & 32.9 & 43.0 & 16.0 \\
OA-Trans\cite{wang2022object}$\clubsuit$ & 18.2 & 34.3 & 43.7 & 18.5 \\
\midrule
RaP(ours)$\clubsuit$  & \textbf{19.7} & \textbf{39.0} & \textbf{47.2} & \textbf{13.0} \\
\bottomrule
\end{tabular}}
\caption{Comparisons with existing text-to-video retrieval state-of-the-art methods with zero-shot and fine-tuning setups on LSMDC.$\clubsuit$: Methods using WebVid2M and CC3M datasets(5.5M). $\diamondsuit$:Methods using HD-VILA-100M\cite{xue2022advancing} dataset.}
\label{table_LSMDC}
\end{table}

\subsection{Experimental Results}
% shot results on the four downstream tasks.
We show results on the four text-video retrieval datasets. 
Since not every previous work has evaluated zero-shot and fine-tuning performance on all datasets, the baseline methods may not be the same for different datasets.
In general, our method achieves significant improvements over the compared methods on all datasets.

\paragraph{MSRVTT  Results} As shown in Table \ref{table_msrvtt}, the zero-shot performance of RaP is 3.0\% higher than the previous best-performed method VIOLET in R1 scores. RaP also achieves the second-highest scores in R5 and R10. 
When fine-tuning on 7k training videos,  RaP outperforms all previous fine-tuned methods and is 4.6\%, 3.3\% higher than ALPRO in R1 and R5. When fine-tuning on 9k training videos,  RaP obtains more than 4.1\% improvement over COTS in R1 scores. 

\paragraph{DiDeMo Results} As shown in Table \ref{table_DiDeMo}, in the zero-shot setting, RaP achieves 5.7\% improvement in R1 over  the previous best-performed method ALPRO.
From the fine-tuning results, RaP outperforms all previous methods. In particular, RaP exceeds ALPRO by +7\% on R1. RaP also reduces the MdR metric of DiDeMo to 2.0, which is smaller than the previous models. 

\paragraph{MSVD Results} Tabel \ref{table_msvd} compares RaP with existing methods on MSVD. The zero-shot performance of RaP surpasses SupportSet and is even slightly higher than FiT's fine-tuning results on R1. After fine-tuning, RaP achieves more than 6.3\%, 6.4\%, and 3.3\% improvement over other fine-tuned models in R1, R5, and R10 scores.

\paragraph{LSMDC Results}  Due to the ambiguity of text descriptions, LSMDC is a more challenging dataset, and the results of previous methods are relatively low. Table \ref{table_LSMDC} shows that RaP outperforms all previous methods in fine-tuning setup, proving RaP's generalization ability in complex scenarios.

From the performance on these different datasets, RaP significantly outperforms the previous methods, which illustrates the importance of reducing the impact of inter-modal redundancy in video-language pre-training. Meanwhile, the improvement also justifies our redundancy measurement and validates the effectiveness of redundancy-aware contrastive learning.

\begin{figure*}[!htbp]
\centering
\includegraphics[width=16cm]{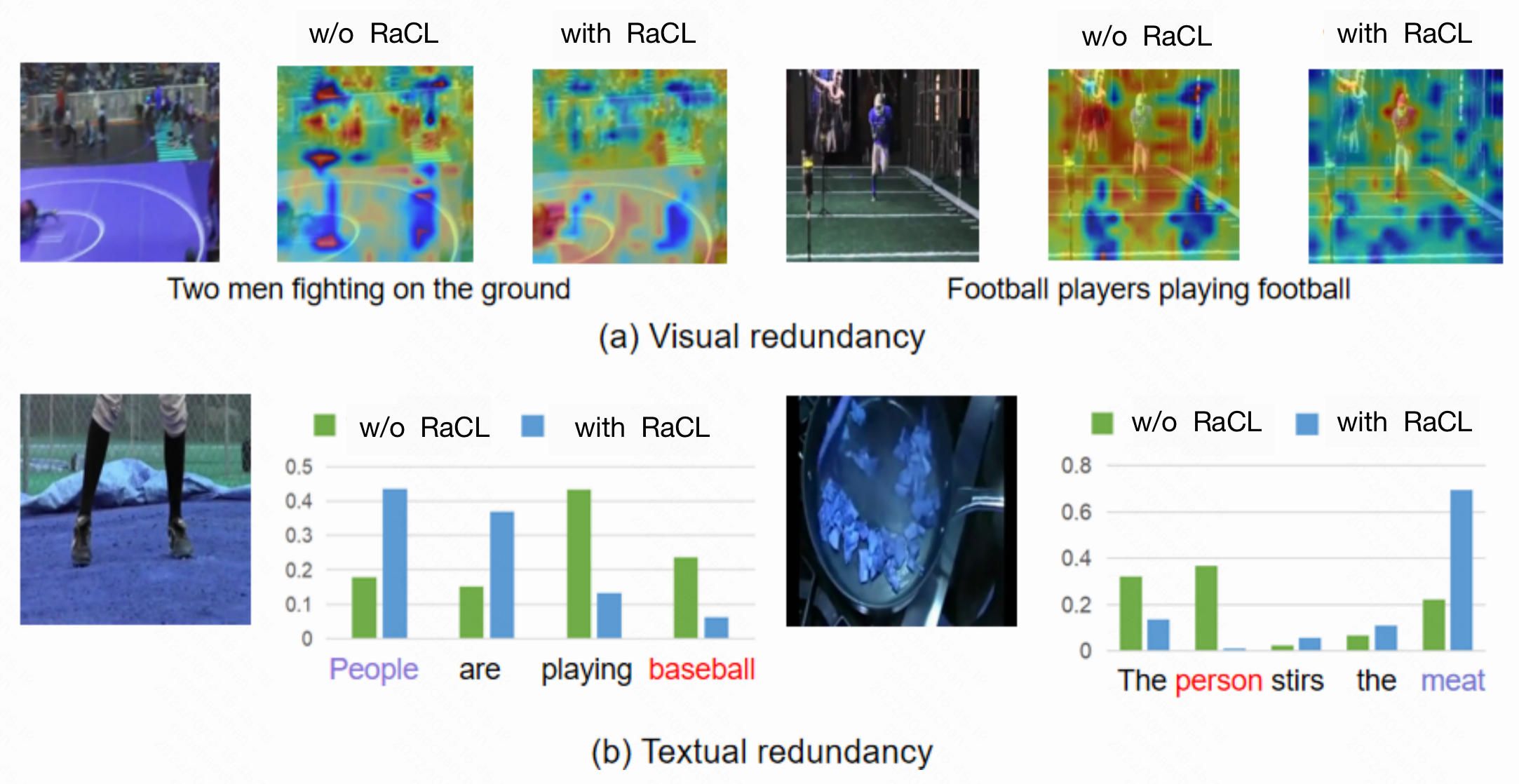}
\caption{Visualization of the effect in identifying redundancy. We train the model without RaCL as our baseline. (a) Visual redundancy: the text [CLS] embedding is a query, and the video non-[CLS] patch embeddings are keys. The darker the blue, the higher the redundancy, the darker the red, the lower the redundancy. (b) Textual redundancy: the video [CLS] embedding is a query and the text non-[CLS] token embeddings are keys. The higher the value, the lower the redundancy.}
\label{guided_case}
\end{figure*}

\begin{table}[!tbp]
\centering
\scalebox{0.7}{
\begin{tabular}{l|cccc}
\toprule
\textbf{Task}  & \textbf{R1}$\uparrow$ & \textbf{R5}$\uparrow$& \textbf{R10}$\uparrow$ & \textbf{MdR}$\downarrow$\\
\midrule
\multicolumn{5}{c}{Zero-shot} \\
\midrule
w/o RaCL  & 24.5 & 45.8 & 54.4 & 8.0  \\
with RaCL$_{v2t}$  & 26.6 & 46.0 & 55.3 & 7.5 \\
with RaCL$_{t2v}$  & 26.3 & 46.5 & 55.1 & 7.0 \\
with RaCL(RaP)  & \textbf{28.9} & \textbf{47.5} & \textbf{56.8} & \textbf{7.0} \\

\bottomrule
\end{tabular}}
\caption{Ablation study of the newly proposed RaCL. We report the results on MSRVTT. The results of the fine-tuning are based on the RaP pre-trained model.}
\label{table_task}
\end{table}

\begin{table}[!t]
\centering
\scalebox{0.7}{
\begin{tabular}{l|cccc}
\toprule 
\textbf{\#frms} &  \textbf{R1}$\uparrow$ & \textbf{R5}$\uparrow$& \textbf{R10}$\uparrow$ & \textbf{MdR}$\downarrow$\\
\midrule
\multicolumn{5}{c}{Zero-shot} \\
\midrule
1  & 21.3 & 40.2 & 49.3 & 11.0 \\
2  & 25.5 & 45.8 & 55.3 & 7.0 \\
4  & \textbf{28.9} & \textbf{47.5} & 56.8 & \textbf{7.0} \\
8  & 26.6 & 45.2 & \textbf{57.5} & 7.0 \\
\bottomrule
\end{tabular}}
\caption{Ablation study of the number of frames.We report zero-shot results on MSRVTT.}
\label{table_frames}
\end{table}

\subsection{Ablations and Analysis}
\label{ablation}
\paragraph{Effect of Redundancy-aware contrastive learning} 
% To learn the effectiveness of the redundancy-aware contrastive learning, we performed two sets of experiments with and without RaP: As shown in Table \ref{table_task}, compared with our base model, adding RaP brings significant improvements.
To further verify the effectiveness of redundancy-aware contrastive learning, we compare the experimental results of the model with and without RaCL on the MSRVTT dataset. As shown in the table \ref{table_task}, either from the results of zero-shot or fine-tuning, removing the RaCL model will decrease performance. Therefore, RaCL plays a crucial role in the pre-training and fine-tuning stages, which is one of the keys that RaP is ahead of others.

\paragraph{Effect of number of frames}
We further explore how using different frame numbers in sparse sampling during pre-training affects the performance of RaP:
We compare sampling 1, 2, 4, and 8 frames from each video, respectively. We report zero-shot results of the model on the MSRVTT dataset. 
As shown in Tabel \ref{table_frames}, when sampling no more than 4 frames, the model's performance gains as the frame number increases. However, when the number of sampled frames increases to 8, the results begin to drop. We believe this is due to the excessive visual redundancy included when sampling more than 8 frames, making it more difficult for the model to learn. Perhaps larger training data can alleviate this problem, which is left to be explored in future work.
% \textbf{Effect of ratio of loss}.
% RaP can be added to other losses without changing the model structure, in order to explore the best ratio configuration, we fixed the existing loss ratio and changed the ratio of RaP: As shown in Table \ref{table_ratio}, when the scale is 1.0, the model can achieve the best performance.
% \textbf{Redundancy-aware contrastive learning on smaller datasets}
% To demonstrate the robustness of our method, we conduct experiments on a smaller WebVid2M dataset. As shown in Table \ref{table_webvid}, Our model outperforms Fit on WebVid2M, illustrating Rap's power on smaller datasets.\\
\paragraph{Effect of coefficient $\lambda$}
In Equation \ref{loss}, a coefficient $\lambda$ is used to balance the RaCL loss in the total loss $\mathcal{L}$. As shown in the Table \ref{table_ratio}, we list the zero-shot results of the model on the MSRVTT dataset when we vary the $\lambda$ values. We find that using $\lambda$ = 1.0 performs the best.

\paragraph{Qualitative Analysis} We provide further visualization for qualitative analysis. 
Specifically, we visualize the weights map between [CLS] embedding from one modal and non-[CLS] embeddings from another, which is a bidirectional process.
Fig \ref{guided_case}-(a) shows the visualization of the weights allocated to each patch. Compared with the scattered concerns of the baseline model, RaP filters out redundant patches and pays more attention to patches that match the text. For example, RaP focuses on the two men fighting in the lower-left corner of the left frame, and the RaP pays attention to the player in the middle of the right frame. The weights of each token are visualized in Fig \ref{guided_case}-(b). Compared with the baseline model, RaP increases the weight of text entities that appear in the frame and decreases the weight of missing entities in the frame. Through RaCL, the [CLS] embedding in RaP can filter redundant information and focus on relevant information between modalities, confirming Rap's significant improvement in the text-video retrieval task.

\begin{table}[!t]
\centering
\scalebox{0.7}{
\begin{tabular}{l|cccc}
\toprule 
\textbf{ratio} &
\textbf{R1}$\uparrow$ & \textbf{R5}$\uparrow$& \textbf{R10}$\uparrow$ & \textbf{MdR}$\downarrow$\\
\midrule
\multicolumn{5}{c}{Zero-shot} \\
\midrule
0.5  & 25.8 & 46.4 & 55.4 & 7.0\\
1.0  & \textbf{28.9} & 47.5 & 56.8 & 7.0 \\
2.0  & 26.7 & \textbf{48.7} & \textbf{57.5} & \textbf{6.0} \\
4.0  & 26.8 & 46.7 & 56.4 & 8.0 \\
\bottomrule
\end{tabular}}
\caption{Ablation study of coefficient $\lambda$. We report zero-shot results on MSRVTT.}
\label{table_ratio}
\end{table}

% \begin{table}[!t]
% \centering
% \scalebox{0.65}{
% \begin{tabular}{l|cccc}
% \toprule 
% \textbf{Method} & \textbf{R1}$\uparrow$ & \textbf{R5}$\uparrow$& \textbf{R10}$\uparrow$ & \textbf{MdR}$\downarrow$ \\
% \midrule
% \multicolumn{5}{c}{Zero-shot} \\
% \midrule
% Divided Space-Time\cite{lei2021less}\dagger & 13.0 & - & 40.2 & 18.0 \\
% FiT\cite{bain2021frozen}\dagger & 14.6 & - & 42.7 & 16.0 \\
% RaP(ours)  & \textbf{18.6} & 35.6 & \textbf{44.1} & \textbf{14.0} \\
% % C  & \textbf{28.9} & \textbf{47.5} & \textbf{56.8} & \textbf{7.0} & \textbf{29.5} & \textbf{55.7} & \textbf{65.6} & \textbf{4.0} \\
% \bottomrule
% \end{tabular}}
% \caption{The models were trained on WebVid-2M. We report the zero-shot results on MSRVTT test set for text-video retrieval. $\dagger$ : results from \cite{bain2021frozen}.}
% \label{table_webvid}
% \end{table}
\section{Conclusion}
In this paper, we summarize the inter-modal redundancy in video language pre-training and propose a redundancy measurement.
Then, we propose redundancy-aware contrastive learning to alleviate redundancy. Significant improvements on several text-video retrieval datasets justify our redundancy measurement and validate the effectiveness of redundancy-aware contrastive learning.

\section{Limitations}
\label{limitation}
Taking the maximum similarity as the weight is a relatively weak constraint. When the redundant information exceeds a specific limit, it may lead to a decrease in the performance of the model. We propose that introducing an attention module to generate weights in future work may improve the model's performance under high redundancy.
% Entries for the entire Anthology, followed by custom entries
\bibliography{anthology,custom}
\bibliographystyle{acl_natbib}

\newpage
\begin{figure*}[!htbp]
\centering
\includegraphics[width=16cm]{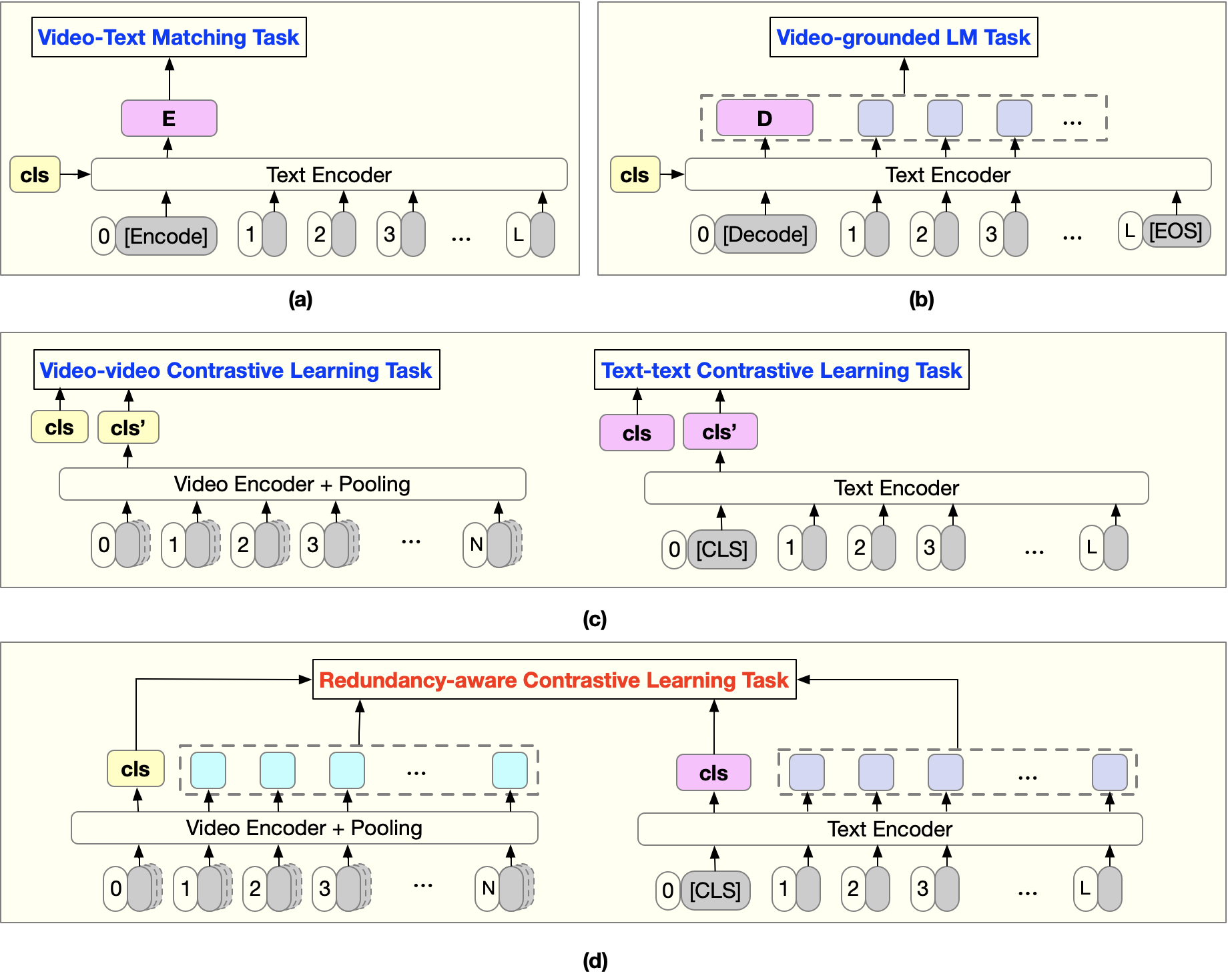}
\caption{The Pre-training tasks. In addition to our designed redundancy-aware contrastive learning task, there are auxiliary tasks: a video-text matching task, a video-grounded language model task and two intra-modal contrastive learning tasks.}
\label{Rap_tasks}
\end{figure*}

\appendix
\section{Appendix}
\label{sec:appendix}
\subsection{Auxiliary Pre-training Tasks}
\paragraph{Video-Text Matching Task} Following \cite{li2022blip}, we inject visual information by inserting an additional cross-attention layer between the self-attention layer and the feed-forward network of each transformer block of the text encoder. 
The modified text encoder shares parameters with the text encoder except for the newly injected layers. As shown in Figure \ref{Rap_tasks}-(a),we construct the input by wrapping the text with a task-specific prompt, i.e. ``[encoded] text''. The output embedding of the additional ``[Encode]'' is the fused multimodal representation of the video and text, which is used to determine whether the video matched the text.

\paragraph{Video-grounded Language Model Task} Following \cite{li2022blip}, we replaces the bidirectional self-attention layers in the video-grounded text encoder with causal self-attention layers.
The modified text encoder shares parameters with the text encoder except for the replaced layers.
As shown in Figure \ref{Rap_tasks}-(b), we construct the input by wrapping the text with a task-specific prompt, i.e. ``[Decode] text''. An end-of-sequence token ``[EOS]'' is used to signal the end of the decoding process. 
\paragraph{Intra-Modal Contrastive Learning Tasks} Following \cite{yang2022vision}, we perform intra-modal contrastive learning on text and video separately to promote a uniform distribution of representations, as shown in Figure \ref{Rap_tasks}-(c). For the video modality, A video will generate two views after data augmentation. We consider these two views as a positive pair. For the text modality, we follow \cite{gao2021simcse} and use standard dropout as a minimal data augmentation method.

These tasks are jointly trained with the redundancy-aware contrastive learning task to optimize the model's parameters together.
\end{document}